УДК 004.032.26

# НЕЙРОСЕТЕВОЙ ПЕРЕНОС ПРОИЗВОЛЬНОГО СТИЛЯ НА ПОРТРЕТНЫЕ ИЗОБРАЖЕНИЯ С ИСПОЛЬЗОВАНИЕМ НЕЙРОСЕТЕЙ С МЕХАНИЗМОМ ВНИМАНИЯ[*]

## С.А. БЕРЕЗИН[1], В.М. ВОЛКОВА[2]

[1]*630073, РФ, г. Новосибирск, пр. Карла Маркса, 20, Новосибирский государственный технический университет, студент кафедры теоретической и прикладной информатики. E-mail: sergeyberezin123@gmail.com*
[2]*630073, РФ, г. Новосибирск, пр. Карла Маркса, 20, Новосибирский государственный технический университет, Доцент кафедры теоретической и прикладной информатики. E-mail: volkova@ami.nstu.ru*

Задача переноса произвольного стиля состоит в создании нового, ранее не существующего изображения, путем комбинирования двух данных изображений: оригинального и стилевого. Оригинальное изображение формирует структуру, основные геометрические линии и формы результирующего изображения, в то время как стилевое изображение задает цвет и текстуру результата. Слово «произвольный» в данном контексте обозначает отсутствие какого-то одного, заранее выученного, стиля. Так, например, свёрточные нейронные сети способные переносить новый стиль только после своего переобучения или дообучения на новом объеме данных, не считаются решающими такую задачу, в то время как сети на основе механизма внимания, способные производить такую трансформацию без переобучения – да. Оригинальное изображение может представлять собою, например, фотографию, а стилевое — картину знаменитого художника. Результирующим изображением в таком случае будет сцена, изображенная на исходной фотографии, выполненная в стилистике данной картины. Современные алгоритмы переноса произвольного стиля позволяют добиться хороших результатов в данной задаче, однако при обработке портретных изображений людей результат работы таких алгоритмов оказывается либо неприемлем, ввиду чрезмерного искажения черт лица, либо слабо выраженным, не носящим характерные черты стилевого изображения. В этой работе рассматрива-ется подход к решению данной проблемы с использованием комбинированной архитектуры глубоких нейронных сетей с механизмом внимания и свёрточной сегментационной сети, осуществляющей перенос стиля с учетом содержимого конкретного сегмента изображения: с ярким преобладанием стиля над формой для фоновой части изображения, и с преобладанием содержания над формой в портретной части изображения, содержащей непосредственно изображение лица человека и/или его фигуры.

**Ключевые слова:** машинное обучение, глубокое обучение, нейронные сети, обработка изображений, перенос стиля, сегментация, свёрточные нейронные сети, механизм внимания.

## ВВЕДЕНИЕ

Методы машинного обучения в общем и нейронные сети в частности, на сегодняшний день находят своё применение в огромном множестве областей [1, 9-15]. Одной из таких областей является обработка изображений. Рассмотрим подробнее задачу переноса стиля, относящуюся к этой области.

Перенос художественного стиля изображения являет собою создание нового изображения, содержащего характерные глобальные и локальные паттерны стилевого изображения, и при этом сохраняющего структуру изображения исходного [1].

Архитектуры на основе свёрточных нейронных сетей успешно справляются с переносом одного [2] или нескольких [3] заранее выученных стилей, однако при выходе за рамки заранее заготовленных шаблонов они требуют полного переобучения.

Ранее предпринимались попытки решить эту задачу с использованием техники адаптивной нормализации данных (adaptive instance normalization) [4], однако субъективная оценка получаемых результатов была далека от идеальной.

С выходом работ за авторством L. Sheng, Z. Lin, J. Shao, X. Wang [6] и Dae Young Park и Kwang Hee Lee [1] в 2018 году, в которых описывается принципиально новый подход к решению этой задачи, произошел значительные рывок в данной области. Предложенные в архитектуре Avatar-Net [6] и доработанные в архитектуре SANet [1] решения с использованием механизма внимания (attention mechanism) позволяют соответствующим образом перестроить характерный паттерн переносимого стиля для каждого участка обрабатываемого изображения с учетом содержания данного участка путем сопоставления отношений, таких как близость (identity loss), между контекстным и стилевым изображениями.

Однако, при обработке портретных изображений любым из вышеприведённых методов неизбежно возникает проблема искажения черт лица: те геометрические преобразования, выученные нейросетью, делающие фон похожим по стилю на стилевое изображение, совершенно искажают форму овала лица, глаз, рта, что приводит к утрате узнаваемости в портрете конкретного человека.

Объектом исследования данной работы стал поиск путей решения данной проблемы, позволяющих сохранить как выразительные геометрические и цветовые преобразования, производимые при переносе стиля, так и узнаваемость черт человеческого лица.

## 1. МЕТОД

Решение, предлагаемое в данной работе, заключается в разбиении исследуемой задачи на две: перенос стиля на фоновую часть исходного изображения с большим весом формы над содержанием, что позволяет сохранить яркие цветовые и геометрические преобразования, и перенос стиля на лицо и фигуру человека с преобладанием веса содержания над формой. Визуализация решения представлена на рисунке 1.

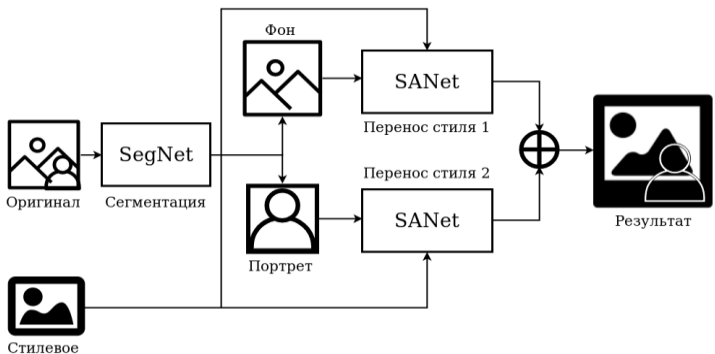

*Рис. 1* – Схема архитектуры решения

Для разделения изображения на две семантические области: фон и фигуру человека, то есть для решения задачи сегментации, была применена нейросеть архитектуры SegNet [5], обученная на наборе данных Microsoft COCO [7].

Выбор именно этой архитектуры обусловлен малыми временными затратами на обработку кадра (около 40 мс на GPU Nvidia Tesla v100) и удовлетворительными, для данной задачи, результатами.

Для переноса стиля использовалась упомянутая ранее архитектура SANet (Style-Attentional Network), которая является передовым решением на текущий момент, обеспечивая наиболее качественный результат [1].

Одной из ключевых особенностей этой модели является использование механизма внимания (attention mechanism) для выявления паттернов характерных для каждого конкретного изображения.

Для варьирования соотношения признаков оригинального изображения к признакам стилевого было предложено ввести два весовых коэффициента. Фактически это осуществлялось путем слияния результатов работы двух слоёв нейросети с разными коэффициентами:

$$Res = sanet4_1 * W_1 + sanet4_2 * W_2,$$

где коэффициенты $W_1$ и $W_2$ определяют степень важности низкоуровневых и высокоуровневых признаков соответственно.

После выделения области с изображением человека составляется бинарная маска, покрывающая эту область. На каждом изображении по полученной маске обрезается фрагмент кадра. Затем, производится первый (фоновый) перенос стиля на оригинальное изображение и второй (портретный) перенос стиля. После этого полученные изображения накладываются друг на друга.

## 2. РЕЗУЛЬТАТЫ ЭКСПЕРИМЕНТОВ

Исходная нейросеть SANet была обучена на наборе данных MS-COCO [7] в качестве исходных изображений и на наборе WikiArt [8] в качестве стилевых изображений. Оба набора содержат приблизительно 80 000 изображений.

Целевой функцией была выбрана функция identity loss, направленной на приоритизацию сохранения структуры изображения, нежели на изменение стилевых характеристик:

$$L_{identity} = \lambda_{identity1}(||I_{cc} - I_c||_2 + ||I_{ss} - I_s||_2)$$
$$+ \lambda_{identity2} \sum_{i=1}^{L} (||\phi_i(I_{cc} - I_c)||_2 + \phi_i(||I_{ss} - I_s||_2))$$

Где $I_{cc}$ и $I_{ss}$ обозначают выходное сгенерированное из двух одинаковых исходных и стилевых изображений, $I_c$ и $I_s$ есть исходное и стилевое изображения, каждый $\phi_i$ обозначает слой нейросети а $\lambda_{identity1}$ и $\lambda_{identity2}$ – гипперпараметры, с экспериментально подобранными значениями 1 и 50 соответственно. Как результат, такая функция потерь позволяет сохранять структуру исходного изображения и стилевые особенности переносимого изображения в одно и то же время [1].

Использовался оптимизатор Adam (adaptive moment estimation) [9] с шагом обучения 0.0001 и размером пакета, равным 5. Применялась аугментация обрезанием части изображений.

На рисунке 2 и 3 представлен поэтапный результат работы предложенной сети.

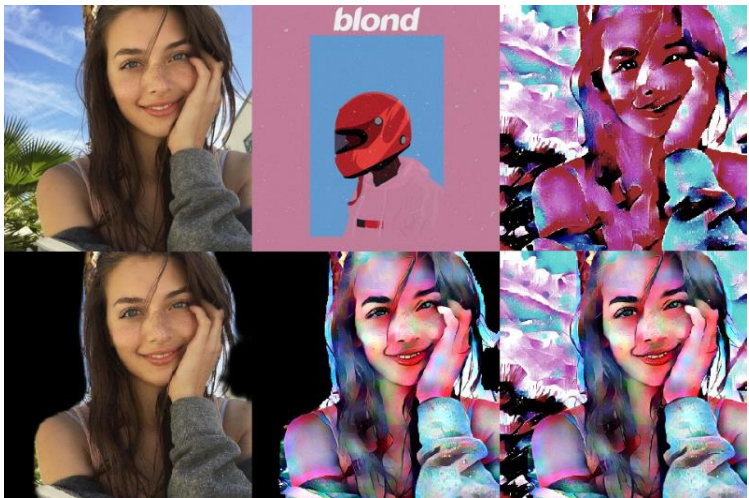

*Рис. 2* – Перенос стиля на портрет. Слева-направо: оригинальное изображение, стилевое изображение, результат прямого переноса стиля, результат сегментации, результат переноса стиля на сегментированное изображение, комбинированный результат.

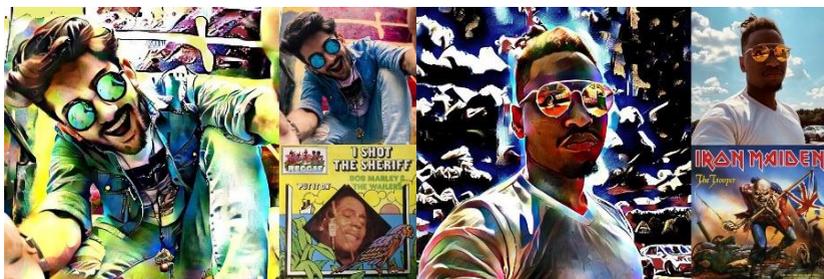

*Рис. 3* – Результат работы алгоритма. Видно соблюдение формы линий человеческого лица при стилизации.

## ЗАКЛЮЧЕНИЕ

Сочетание всех вышеприведённых методов и техник позволило добиться улучшения качества переноса стиля с произвольного изображения на человеческие портреты в сравнении с исходными подходами. Найденное решение было внедрено в конечный программный комплекс.

***Березин Сергей Андреевич***, студент кафедры теоретической и прикладной информатики Новосибирского государственного технического университета, техник лаборатории бизнес-решений на основе искусственного интеллекта Московского физико-технического института. Основное направление научных исследований – технологии машинного обучения. E-mail: sergeyberezin123@gmail.com

***Волкова Виктория Михайловна***, доцент кафедры теоретической и прикладной физики Новосибирского государственного технического университета, кандидат технических наук. Основное направление научных исследований машинное обучение, анализ данных. Имеет более 30 публикаций. E-mail: volkova@ami.nstu.ru

## Neural arbitrary style transfer for portrait images using the attention mechanism[*]

### S. A. Berezin[1], V.M. Volkova[2]

[1]*Novosibirsk State Technical University, 20 Karl Marks Avenue, Novosibirsk, 630073, Russian Federation, undergraduate of the department of theoretical and applied informatics.*
*E-mail: sergeyberezin123@gmail.com*
[2]*Novosibirsk State Technical University, 20 Karl Marks Avenue, Novosibirsk, 630073, Russian Federation, PhD in engineering, associate professor of the department of theoretical and applied informatics.*
*E-mail: volkova@ami.nstu.ru*

Arbitrary style transfer is the task of synthesis of an image that has never been seen before, using two given images: content image and style image. The content image forms the structure, the basic geometric lines and shapes of the resulting image, while the style image sets the color and texture of the result. The word "arbitrary" in this context means the absence of any one pre-learned style. So, for example, convolutional neural networks capable of transferring a new style only after training or retraining on a new amount of data are not considered to solve such a problem, while networks based on the attention mechanism that are capable of performing such a transformation without retraining - yes. An original image can be, for example, a photograph, and a style image can be a painting of a famous artist. The resulting image in this case will be the scene depicted in the original photograph, made in the stylie of this picture. Recent arbitrary style transfer algorithms make it possible to achieve good results in this task, however, in processing portrait images of people, the result of such algorithms is either unacceptable due to excessive distortion of facial features, or weakly expressed, not bearing the characteristic features of a style image. In this paper, we consider an approach to solving this problem using the combined architecture of deep neural networks with a attention mechanism that transfers style based on the contents of a particular image segment: with a clear predominance of style over the form for the background part of the image, and with the prevalence of content over the form in the image part containing directly the image of a person.

---

[*]*Manuscript received on December 20. 2019*

**Keywords:** machine learning, deep learning, neural networks, image processing, style transfer, segmentation, convolutional neural networks, attention mechanism.

***Berezin Sergey Andreevich***, undergraduate of the Department of Theoretical and Applied Informatics, Novosibirsk State Technical University, technician of the laboratory of business solutions based on artificial intelligence of the Moscow Institute of Physics and Technology. The main direction of scientific research is machine learning technology. Email: sergeyberezin123@gmail.com

***Viktoriya M. Volkova.*** Novosibisrk State Technical University, department of theoretical and applied informatics, volkova@ami.nstu.ru, PhD in engineering, associate professor. Main areas of scientific interest: machine learn-ing, data science, investigations of statistical criteria under conditions of basic assumptions violation. She is the author coauthor of more of over 30 research papers and teaching manuals.